\documentclass[conference]{IEEEtran}
\IEEEoverridecommandlockouts
\usepackage{cite}
\usepackage{amsmath,amssymb,amsfonts}
\usepackage{graphicx}
\usepackage{textcomp}
\usepackage{xcolor}
\usepackage{mathtools} %
\usepackage{todonotes} %
\usepackage[caption=false]{subfig} %
\usepackage[mathscr]{euscript} 
\usepackage{url} %
\usepackage{hyperref}
\usepackage{bm}
\usepackage{algpseudocode} 
\usepackage{algorithm}
\usepackage{xifthen}
\usepackage[export]{adjustbox}
\usepackage{amsmath}
\usepackage{xcolor}
\usepackage{listings}

\definecolor{mygreen}{rgb}{0,0.6,0}
\definecolor{mygray}{rgb}{0.5,0.5,0.5}
\definecolor{mymauve}{rgb}{0.58,0,0.82}

\lstdefinelanguage{cypher}{
    sensitive=true,
    morekeywords=[1]{MATCH, RETURN, WHERE},
    morekeywords=[2]{PERSON, FRIEND},
    morestring=[b]",
    morecomment=[l]{//},
    morecomment=[s]{/*}{*/},
    morecomment=[s]{--}{\ },
}

\lstdefinestyle{cypherstyle}{
    language=cypher,
    basicstyle=\small\ttfamily,
    keywordstyle=[1]\color{blue},
    keywordstyle=[2]\color{red},
    commentstyle=\color{mygreen},
    stringstyle=\color{mymauve},
    numberstyle=\tiny\color{mygray},
    breaklines=true,
    showstringspaces=false,
    captionpos=b
}

\usepackage{float}
\usepackage{subcaption} 
\usepackage{svg}
\newcommand{\mycomment}[1]{}
\NewDocumentCommand{\vect}{ O{} O{} m }{\mathbf{#3}\ifthenelse{\isempty{#1}}{}{^{(#1)}}\ifthenelse{\isempty{#2}}{}{_{#2}}}

\NewDocumentCommand{\mat}{ O{} O{} m }{\mathbf{#3}\ifthenelse{\isempty{#1}}{}{^{(#1)}}\ifthenelse{\isempty{#2}}{}{_{#2}}}

\NewDocumentCommand{\ten}{ O{} O{} m }{\pmb{\mathscr{#3}}\ifthenelse{\isempty{#1}}{}{^{(#1)}}\ifthenelse{\isempty{#2}}{}{_{#2}}}

\def\BibTeX{{\rm B\kern-.05em{\sc i\kern-.025em b}\kern-.08em
    T\kern-.1667em\lower.7ex\hbox{E}\kern-.125emX}}
\begin{document}
\title{ Cyber-Security Knowledge Graph Generation by Hierarchical Nonnegative Matrix Factorization}

\author{\IEEEauthorblockN{
Ryan Barron\IEEEauthorrefmark{2}\IEEEauthorrefmark{3}, Maksim E. Eren\IEEEauthorrefmark{1}\IEEEauthorrefmark{3}, Manish Bhattarai\IEEEauthorrefmark{2},
 Selma Wanna\IEEEauthorrefmark{1},\\ Nicholas Solovyev\IEEEauthorrefmark{2}, Kim Rasmussen\IEEEauthorrefmark{2}, 
Boian S. Alexandrov\IEEEauthorrefmark{2},Charles Nicholas \IEEEauthorrefmark{3} and Cynthia Matuszek\IEEEauthorrefmark{3}}
\IEEEauthorblockA{
\IEEEauthorrefmark{2}Theoretical Division, Los Alamos National Laboratory. Los Alamos, USA. \\
\IEEEauthorrefmark{1}Advanced Research in Cyber Systems, Los Alamos National Laboratory. Los Alamos, USA. \\
\IEEEauthorrefmark{3}Department of Computer Science and Electrical Engineering, University of Maryland, Baltimore County. Maryland, USA.}
\thanks{U.S. Government work not protected by U.S. copyright.}
}

\maketitle
\vspace{-40em}

\begin{abstract}

Much of human knowledge in cybersecurity is encapsulated within the ever-growing volume of scientific papers. As this textual data continues to expand, the importance of document organization methods becomes increasingly crucial for extracting actionable insights hidden within large text datasets. Knowledge Graphs (KGs) serve as a means to store factual information in a structured manner, providing explicit, interpretable knowledge that includes domain-specific information from the cybersecurity scientific literature. One of the challenges in constructing a KG from scientific literature is the extraction of ontology from unstructured text. In this 
paper, we address this topic and introduce a method for building a multi-modal KG by extracting structured ontology from scientific papers. We demonstrate this concept in the cybersecurity domain. One modality of the KG represents observable information from the papers, such as the categories in which they were published or the authors. The second modality uncovers latent (hidden) patterns of text extracted through hierarchical and semantic non-negative matrix factorization (NMF), such as named entities, topics or clusters, and keywords. We illustrate this concept by consolidating more than two million scientific papers uploaded to arXiv into the cyber-domain, using hierarchical and semantic NMF, and by building a cyber-domain-specific KG.

\end{abstract}

 



\maketitle

\section{Introduction}
\label{sec:introduction}

With the ever-growing volumes of text data, organizing large quantities of scientific papers and identifying patterns within the scientific community continues to be a challenging problem. Knowledge Graphs (KGs) are one of the key techniques for storing factual knowledge in a structured manner. However, building KGs from scientific literature remains a challenge since much of the useful context in data, such as entities, relationships between entities, and properties of entities, is hidden in an unstructured manner. At the same time, domains such as cybersecurity are highly diverse, containing knowledge from a multidisciplinary set of fields. This necessitates a more granular or highly domain-specific organization to discover sub-topic areas. In this paper, we introduce a concept for building domain-specific, multi-modal KGs using hierarchical and semantic Non-negative Matrix Factorization (NMF) for topic modeling. In our concept framework, we consolidate generic scientific corpora into a highly-specific domain and use the extracted latent (hidden) patterns as an ontology in the KG.

Here, for extracting highly-domain-specific clusters, we introduce a hierarchical approach named HSNMFk-SPLIT that can extract fine-grained sub-topics and their semantic sub-structures from large text corpora, using NMF with automatic model determination (NMFk \cite{TELF}). For semantic NMFk, given a text-document matrix $\mat{X} \in \rm I\!R_{+}^{F \times N}$ where $F$ is the number of tokens in the vocabulary and $N$ is the number of documents and a word-context (co-occurrence) matrix $\mat{S} \in \rm I\!R_{+}^{F \times F}$, the number of hidden topics, $k$, can be automatically estimated through joint factorization. Here we enact joint factorization with the SPLIT method, which allows combination of patterns from separate factorizations. Incorporating the semantic structure of the text with the ability to estimate the number of topics enables a coherent separation of the latent topics and accurate document clustering \cite{9521777, erenDocEng22}. We apply the SPLIT method in a hierarchical manner to expand and separate the main topics into sub-topics for a chosen domain, where the extracted sub-semantic structures serve as narrow vocabularies or scientific-jargon seeds for Name Entities Recognition (NER). Furthermore, to enhance the semantic clustering in each topic, we also jointly factorize, using SPLIT, the category-text matrix $\mat{C} \in \rm I\!R_{+}^{F \times L}$ ($L$ is the number of distinct categories), values of which represent the TF-IDF of tokens per document category. Finally, HSNMFk-SPLIT addresses the computation overhead that arises from factorizing a large TF-IDF matrix $\mat{X}$ by separating $N$ documents into an arbitrary number of distinct matrices, factorizing each smaller TF-IDF matrix separately in a distributed manner \cite{TELF}, and finally combining the patterns from each matrix with the SPLIT method.

We demonstrate our hierarchical method by performing topic modeling of titles and abstracts from papers posted on arXiv (\url{{https://arxiv.org/}}), which number more than two million. We then select a main topic of Deep Learning, followed by a sub-topic of Security, and then a sub-sub topic of Cybersecurity and Adversarial ML to build a domain specific KG of observable data (titles, authors, etc.) and the latent patterns (NER, keywords, and clusters/topics). This KG can be used to identify emerging trends and facilitate the discovery of relevant research papers for cybersecurity professionals in highly specific domains such as Adversarial ML. Our initial results when demonstrating this concept show the ability and practicality of our HSNMFk-SPLIT to extract meaningful information from topics and their semantic sub-structures from large datasets.


\section{Relevant Work}
\label{sec:relevant_work}

\subsection{Knowledge Graphs}
Recent studies have demonstrated the application of knowledge graphs in identifying emerging trends, uncovering hidden connections, and predicting future research directions within scientific domains \cite{manghi2021new, pingle2019relext,luan-etal-2018-multi,WANG2018112, textbook_graph, SHAO2021113764, ABUSALIH2021103076, KNOWEDU}. The integration of semantic analysis with graph-based structures allows for a more nuanced understanding of the thematic evolution in fields such as deep learning and cybersecurity \cite{opdahl2022semantic, sikos2023cybersecurity}. These advances highlight the potential of knowledge graphs to transform the exploration and exploitation of scientific literature, moving beyond traditional search mechanisms to enable a deeper, contextually rich engagement with content \cite{auer2020improving, zhu2017use}.
To achieve these results, care must be given to the specific process for constructing and analyzing knowledge graphs, as studied and explained in \cite{cyber_ontology,CHEN2019100959, k12_edu_graph, ACEKG, Knowledgegraph_theory}.

Similar to our work, \cite{Auto_web_scraping}, scrapes the web for information relevant to the domain specific knowledge. However their domain is in medicine, and ours is in cyber security. 
Further, they do not use decompositions to construct the graph, but rather vocabularies. Similarly,  \cite{covid_graph}, uses dates, topics, and events as entities to make the knowledge graph but omits the use of decomposition to organize large data.
Our decompositions are enhanced by incorporating the semantic structure of the text with the ability to estimate the number of topics, which enables a coherent separation of the latent topics and accurate document clustering. This method is supported by the findings in \cite{9521777, erenDocEng22}. Building on \cite{9521777, erenDocEng22}, we expand our approach through hierarchical decomposition, as well as incorporating semi-supervised labels of arXiv categories in joint factorization to refine our capabilities.



\subsection{Hierarchical Topic Modeling}

A common method to hierarchical topic modeling is through a probabilistic approach, in which the corpus is treated as a mixture of topics that can be modeled with a probability distribution. One such method is hierarchical Latent Dirichlet Allocation (hLDA) using the nested Chinese Restaurant Process (nCRP) beforehand \cite{blei2003ncrp}. This model allows for arbitrarily large branching factors in topic hierarchies with the nCRP aiding in inferring the appropriate depth of the topic tree. Another LDA-based approach is Hierarchical Dirichlet Processes (HDP). In this method, a set of Dirichlet processes are coupled via their base measure, which is itself distributed according to a Dirichlet process \cite{Teh2006HierarchicalDP}. The authors introduce two Markov Chain Monte Carlo (MCMC) sampling schemes for posterior inference under hierarchical Dirichlet process mixtures and test the approaches on multiple datasets. 


More recent approaches to hierarchical topic modeling make use of variational auto-encoders, recurrent neural networks, and transformers \cite{jin2021neural, guo2020recurrent, Viegas2020CluHTMS, Mifrah_Benlahmar_2022, grootendorst2022bertopic}. While successful, these methods face a problem common to deep learning: successful models depend heavily on the quantity and quality of the training data. HSNMFk-SPLIT is a favorable approach to hierarchical topic modeling since it is instance-based. Only the data that is being modeled is required for computation.



\section{Dataset and Pre-processing}
\label{sec:dataset}

Using HSNMFk-SPLIT, we perform topic modeling on the titles and abstracts of literature uploaded to arXiv\footnote{arXiv Dataset: \url{https://www.kaggle.com/datasets/Cornell-University/arxiv}} \cite{clement2019arxiv}. Our pre-processing pipeline includes removal of common English stop-words, stop-words that are often present in scientific literature (such as `doi', `preprint', `copyright', `figure', and `demonstrate'), stop-phrases (such as `All rights reserved.'), non-ASCII characters, symbols, next-line characters, tags (such as HTML and LaTeX), and e-mail addresses. We also join hyphenated words into single tokens, make the tokens lower-case, and lemmatize the tokens using the Python package \textit{NLTK} \cite{bird2009natural} (NLTK is used with the allowed postags `NOUN', `ADJ', `VERB', `ADV', and `PROPN'). The non-English abstracts are excluded from analysis using the Python implementation of \textit{language-detection} software \cite{nakatani2010langdetect}. Finally, after removing those abstracts with less than 10 tokens, we are left with 2,178,187 documents in our corpus represented with a TF-IDF matrix $\mat{X} \in \rm I\!R_{+}^{10,280 \times 2,178,187}$. To reduce noise, we remove the tokens present in more than 80\% and less than 500 documents when building $\mat{X}$.

The semantic structure of the documents is represented with matrix $\mat{S} \in \rm I\!R_{+}^{10,280 \times 10,280}$ where the values represent the number of times two words co-occur in a predetermined window length of $w=100$ tokens. We normalize $\mat{S}$ with Shifted Positive Point-wise Mutual Information (SPPMI) \cite{levy2014neural}, with shift $s=4$. The categorical semantic structure is represented with the matrix $\mat{C} \in \rm I\!R_{+}^{10,280\times 172}$ values of which represents the TF-IDF of tokens per document category. Here the category of the document is assigned by the author in arXiv\footnote{arXiv Category Taxonomy: \url{{https://arxiv.org/category\_taxonomy}}}. While a given document can be assigned to more than one category, we use the primary category assignment.

\section{Methods}
\label{sec:methods}
\begin{figure*}[t!]
\centering
\includegraphics[width=.9\linewidth]{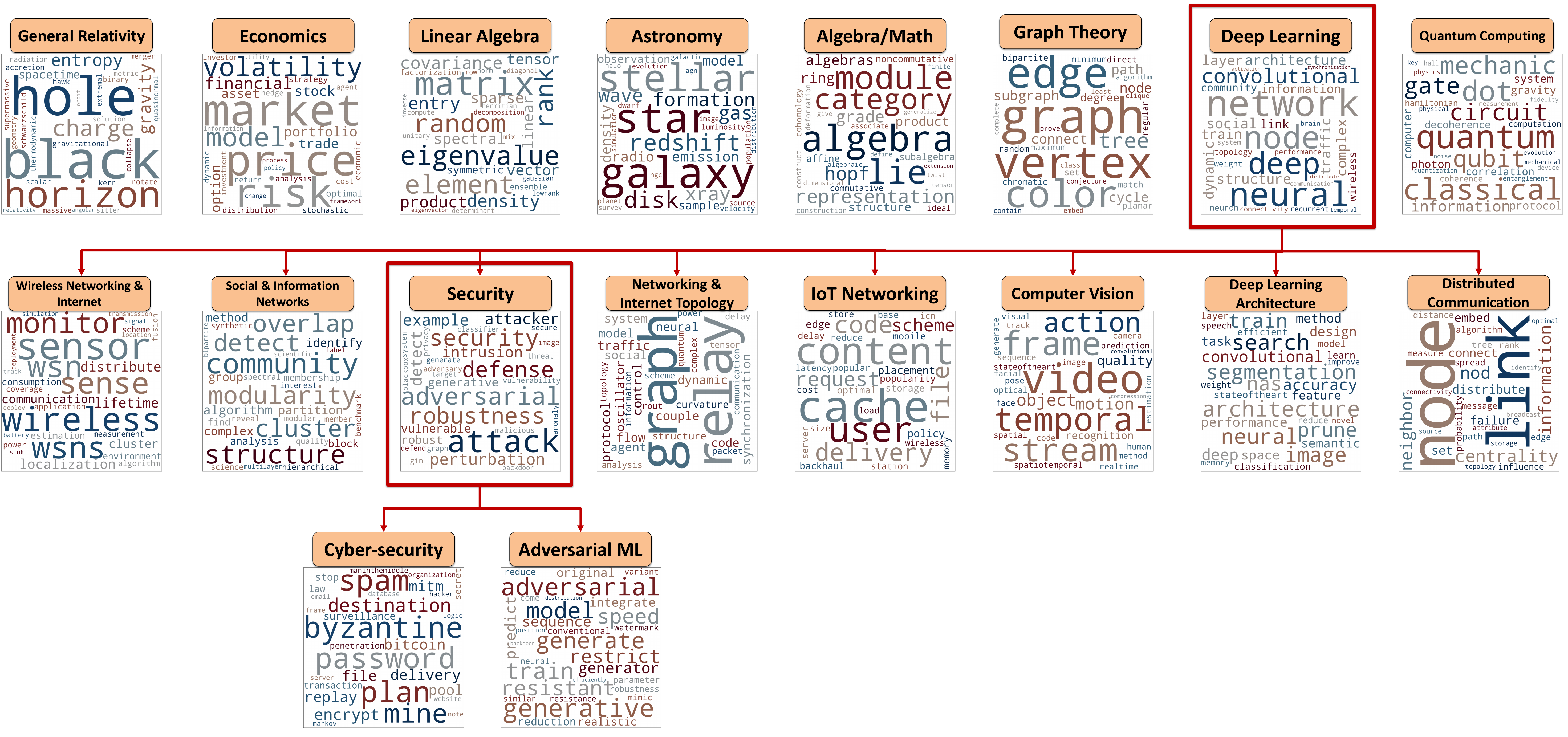}
\caption{Word cloud of selected topics and their interpreted labels. Method is hierarchically applied three times, selecting a topic to extract its sub-topics at each level. Selected topics from each level are shown in each row of the figure. \label{fig:topics}}
\vspace{-1em}
\end{figure*}

In this section, we first describe the HSNMFk-SPLIT method, then explain how we use the clusters found by HSNMFk-SPLIT to build our KG.

\subsection{HSNMFk-SPLIT}

We use a publicly available implementation of NMFk for its automatic model determination capability, which avoids over/under-fitting to topics, and apply it in a hierarchical manner \cite{TELF}\footnote{NMFk is available in \url{https://github.com/lanl/T-ELF}.}. HSNMFk-SPLIT is a method that automatically estimates the number of latent topics and extracts coherent topics by exploiting the semantic representation encoded in the word-context matrix, categorical information encoded in the word-category matrix, together with the text-document matrix. Given a TF-IDF text-document matrix $\mat{X}\in \mathbb{R}_{+}^{F\times N}$, SPPMI normalized word-context matrix $\mat{S}\in \mathbb{R}_{+}^{F\times F}$, and word-category matrix $\mat{C}\in \mathbb{R}_{+}^{F\times L}$ at each node of the hierarchical tree, HSNMFk extracts the topics - the columns of matrix $\mat{W}$ and coordinates of the documents, i.e., columns of matrix $\mat{H}$ corresponding to the $\mat{X}$ for that node in the tree. HSNMFk is performed by solving a joint optimization problem at each node in the tree:

\begin{equation}
\begin{aligned}
    \underset{\mat{W} \in \mathbb{R}_{+}^{F\times k}, \mat{H} \in \mathbb{R}_{+}^{k \times N}, \mat{G} \in \mathbb{R}_{+}^{k \times F}, \mat{J} \in \mathbb{R}_{+}^{k \times L}}{\operatorname{minimize}} \\ 
 \frac{1}{2} ||\mat{X} - \mat{WH}||_F^2 +\alpha ||\mat{S} - \mat{WG}||_F^2+\\
\beta ||\mat{C} - \mat{WJ}||_F^2
\end{aligned}
\end{equation}

where $||.||_F^2$ is the Frobenius norm, and $\alpha$ and $\beta$ are regularization parameters controlling the weight of $\mat{S}$ and $\mat{C}$ in the decomposition. $F$ is the number of words in the vocabulary, $N$ is the number of documents, and $L$ is the number of categories. One way of evaluating the above expression is by concatenating the TF-IDF matrix $\mat{X}$ with $\mat{S}$, and $\mat{C}$, and finally applying NMFk \cite{TELF} on the concatenation \cite{stanev2021topic}. 
However, this is computationally challenging as the matrices are significantly larger after concatenation. Hence, we incorporate the SPLIT technique to minimize the computation and be able to perform the decomposition of each matrix in parallel (HSNMFk-SPLIT):
\vspace{-1.0em}
\\
\begin{itemize}
    \item \textbf{Factorizing large matrices via SPLIT:} First, split $\mat{X}\in \mathbb{R}_{+}^{F\times N}$ into $m$ chunks so that the $i^{th}$ chunk given by $\mat[][i]{X}\in \mathbb{R}_{+}^{F\times \frac{N}{m}}$ can be factorized with NMFk as $\mat[][i]{X}\approx \mat[][i]{W}\mat[][i]{H}$ where $\mat[][i]{W}\in \mathbb{R}_{+}^{F\times k_i}$ and $\mat[][i]{H}\in \mathbb{R}_{+}^{k_i\times \frac{N}{m}}$. Here $m$ chunks can be factorized in parallel. 
    Next, concatenate  $m$ $\mat[][i]{W}$ matrices obtained from previous step to construct $\tilde{\mat{W}}\in \mathbb{R}_{+}^{F\times K}$ where $\tilde{\mat{W}}$=$[\mat[][1]{W}|\mat[][2]{W}|...|\mat[][m]{W}]$ and $K=k_1+k_2+.....+k_m$. Now factorize $\tilde{\mat{W}}$ with NMFk such that $\tilde{\mat{W}} \approx \mat[][x]{W}\mat[][x]{M}$ where $\mat[][x]{W}\in \mathbb{R}_{+}^{F\times p}$,$\mat[][x]{M}\in \mathbb{R}_{+}^{p\times K}$ and $\mat[][x]{M}=[\mat[][1]{M}|\mat[][2]{M}|...|\mat[][m]{M}]$ where $\mat[][j]{M}\in \mathbb{R}_{+}^{p\times k_i}$.
    Finally, from previous steps, multiply $\mat[][i]{M}$ and $\mat[][i]{H}$ to obtain $\mat[*][i]{H}$ where  $\mat[*][i]{H}$=$\mat[][i]{M} \times \mat[][i]{H}$ and $\mat[*][i]{H}\in \mathbb{R}_{+}^{p\times \frac{M}{m}}$ 

    \item \textbf{Incorporate semantic and category structure:} Given word SPPMI matrix $\mat{S}$ and category matrix $\mat{C}$, utilize NMFk to decompose  $\mat{S}\approx \mat[][s]{W} \mat[][s]{H}$ and $\mat{C}\approx \mat[][c]{W} \mat[][c]{H}$ where $\mat[][s]{W}\in \mathbb{R}_{+}^{F\times s}, \mat[][s]{H}\in \mathbb{R}_{+}^{s\times F},\mat[][c]{W}\in \mathbb{R}_{+}^{F\times c}$ and $ \mat[][c]{H}\in \mathbb{R}_{+}^{c\times L}$. Now concatenate $\mat[][x]{W}$ with $\mat[][s]{W}$ and  $\mat[][c]{W}$ to obtain $\mat[][+]{W}$ where $\mat[][+]{W}=[\mat[][x]{W}|\mat[][s]{W}|\mat[][c]{W}]$ and $\mat[][+]{W}\in \mathbb{R}_{+}^{F\times Z}$ where $Z=p+s+c$. Then factorize $\mat[][+]{W}$ with NMFK as $\mat[][+]{W}=\mat[][]{W}\mat[][]{Y}$ where $\mat[]{W}\in \mathbb{R}_{+}^{F\times t}$, $\mat[]{Y}\in \mathbb{R}_{+}^{t\times Z}$, $\mat[]{Y}= [\mat[][x]{Y}|\mat[][s]{Y}|\mat[][c]{Y}]$ and $\mat[][x]{Y}\in \mathbb{R}_{+}^{t\times p}, \mat[][s]{Y}\in   \mathbb{R}_{+}^{t\times s}$ and $\mat[][c]{Y}\in   \mathbb{R}_{+}^{t\times c}$. 
    Finally, multiply $\mat[][x]{Y}$ with $\mat[*][]{H}$ to obtain $\mat{H}$ such that  $\mat{H}=\mat[][x]{Y}\mat[*][]{H}$ where $\mat{H}\in \mathbb{R}^{t\times N}_{+}$ and $\mat{W}\in \mathbb{R}^{F\times t}_{+}$

    \item \textbf{Hierarchical topic modeling:} For each node in the tree, once $\mat{W}$ and $\mat{H}$ are obtained, $\mat{W}$ models the topics, and $\mat{H}$ models the document's topic assignment. Here we compute the maximum indices for each column of $\mat{H}$ (also known as H-clustering)\cite{vangara2021finding} to estimate the best topic representation of each document given as topic($\mat[][j]{X}$) = ${\operatorname{argmax} \mat[][j]{H}}$. The documents in each topic (super-topic) can then be factorized again following the above steps to estimate the underlying sub-topics along the tree depth.
 \end{itemize}

\subsection{Building the Knowledge Graph}
\begin{figure*}
    \centering
 
 \includegraphics[width=.85\textwidth]{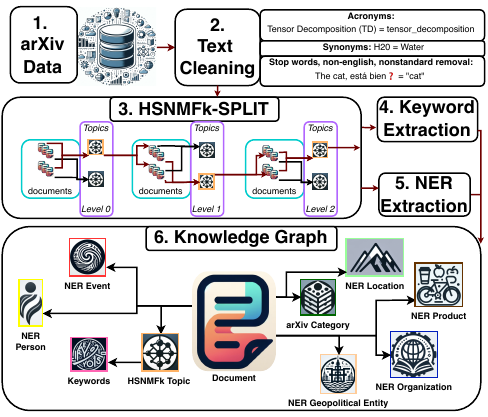}
 
    \caption{Data pipeline from starting from arXiv data (1), text cleaning (2),  running HSNMFk-SPLIT for topics (3), extracting the keywords for the decomposition (4), extracting the Named Entities per document (5), and then structurally aggregating the data into the knowledge graph (6). Images generated with DALL·E \cite{dalle_tensor_decomp_arxiv_images}. }
    \label{fig:kg}
\end{figure*}

The knowledge graph, a Neo4j graph\cite{neo4j2023 }, is composed of nodes and relations from several forms of data . First and foremost, the documents are inserted into the graph, where they are assigned observable attributes in the original data. These observable data include titles, authors, DOI, year, publication information, citations, references, affiliations, etc. Next, the cleaned text decompositions are used to extract HSNMFk-SPLIT topics and keywords from a highly domain-specific leaf decomposition about security. The words and topics are injected into the graph, connected to their documents from the $arg max$ of the topic probability for each document. The keywords are connected to the topics where the edges contain the probabilities as weights. Next, the text of each document is examined with spaCy's NER pipeline \cite{spacyEnCoreWebTrf }, which produces 18 labels. We insert the 6 most useful labels of the 18 total for the purpose of cyber-security ontological exploration:Organization, Event, Person, Location, Product, and Geopolitical Entity. Each of the NERs is connected to each source document that produces it; some of documents may produce the same NER entities. Documents are further connected into communities where  NERs are shared.


\section{Results}
\label{sec:results}

\begin{figure}[htb]
\centering
\includegraphics[width=\linewidth]{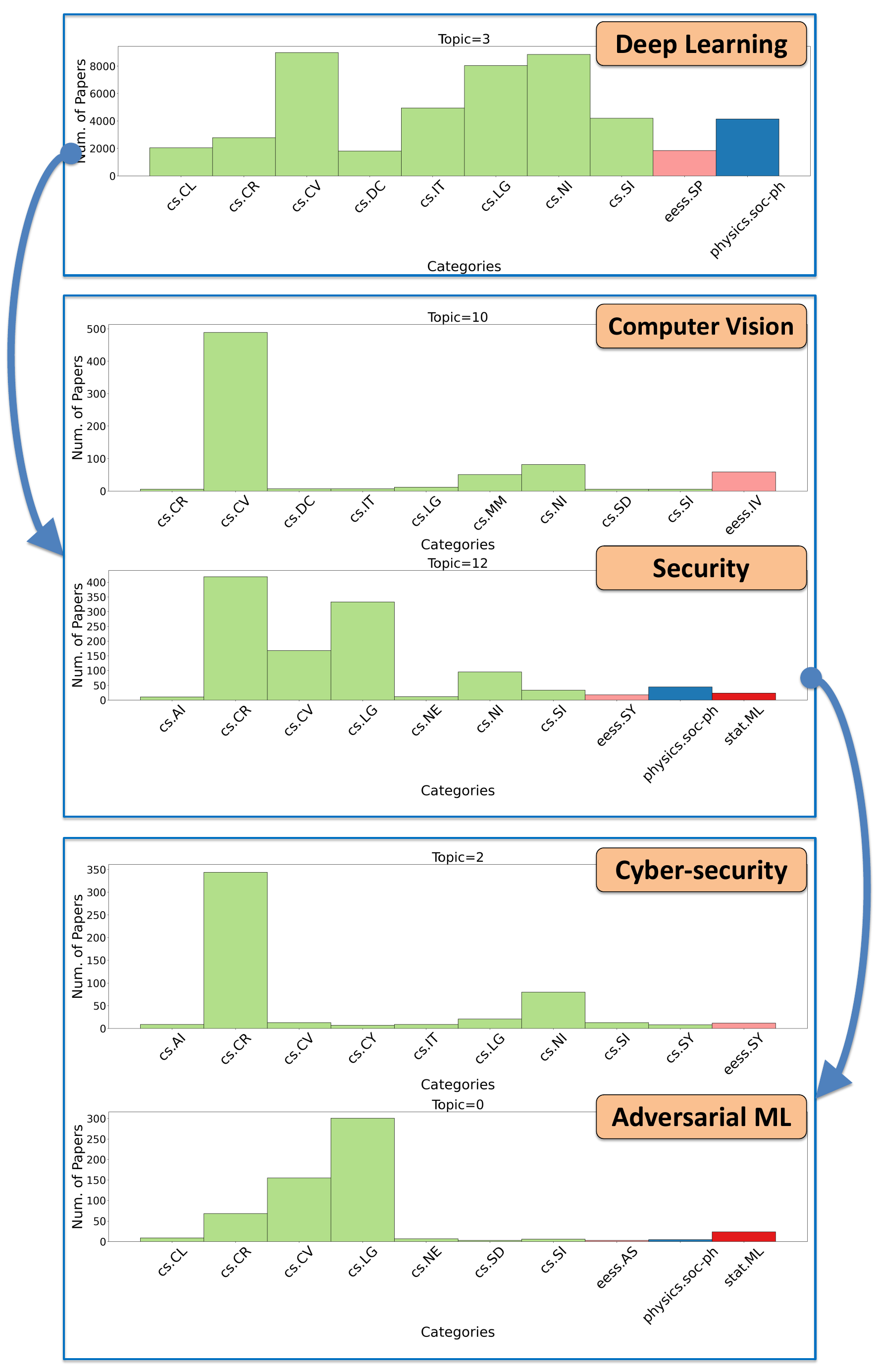}
\caption{Distribution of the top ten document categories, based on the arXiv author assignments. Each topic corresponds to the word clouds from Figure \ref{fig:topics}. Selected topics from each level are shown in each row of the figure. \label{fig:categories}}
\vspace{-1em}
\end{figure}

Our results are shown in Figure \ref{fig:topics} with the word clouds (most prominent words corresponding to each column of $\mat{W}$), and with the distribution of categories in each topic and sub-topic in Figure \ref{fig:categories} (the top ten categories are shown for the given topic). While word clouds highlight the interpretability and quality of the extracted topics, the distribution of the categories shows the specificity of the documents as we look at the sub-categories. We apply our method hierarchically three times, at each level selecting one topic to expand further to identify its sub-topics. Since the initial $\mat{X}$ is large, we chunk the documents into 20 separate matrices and apply semantic joint factorization with automatic determination of the number of latent topics. Individually, the 20 matrices revealed between 30 and 80 latent topics. After joint factorization, we have identified 24 total super-topics in 2 million+ documents. 

We selected the topic related to Deep Learning. The first row of Figure \ref{fig:topics} shows the word cloud for this topic, and the first row of Figure \ref{fig:categories} displays the distribution of categories that are in this topic. The categories are mainly represented in diverse sub-fields of Computer Science, ranging from Cyber (\textit{cs.CR}) to Computer Vision (\textit{cs.CV}). At the second level, the sub-topic of Deep Learning produced 29 total sub-topics, which include Computer Vision and Security. The sub-topic that can be interpreted as Computer Vision based on the tokens in the word cloud is now narrowed down to the \textit{cs.CV}. On the other hand, the Security sub-topic is mainly distributed between the categories \textit{cs.CR} (Cyber), \textit{cs.CV} (Computer Vision), and \textit{cs.LG} (Language); therefore, we expand the Security sub-topic to the third level, further separating it into three sub-sub-topics, two of which can be interpreted as Cyber-security and Adversarial ML based on the word clouds. Similarly at the third level, the Cyber-security topic category distribution is narrowed down to \textit{cs.CR}. The Adversarial ML topic contains documents from three main categories of \textit{cs.CR}, \textit{cs.CV}, and \textit{cs.LG}, which is reasonable based on this field and how research focuses on adversarial computer vision and language models.


In the domain specific leaf decomposition, 1383 vocabulary words were used to discover three latent topics, which are the counts of the entities in the graph. These three topics contain their own unique, non-overlapping set of documents. Specifically, topic 0 had 597 documents, topic 1 had 114 documents, and topic 2 had 565 documents. These topics were composed from the 1383 vocabulary words, but the probability of the words occurring in each topic changes according to the documents contained. The knowledge graph contained NER entities such as 245 organizations, 7 events, 64 persons, 3 locations, 679 products, and 22 geopolitical entities. Overall the graph produced 3758 node entities and 9428 edge relationships. Queries could be asked of the knowledge graph, which have to first be translated from natural language into Cypher language. For example, 'Which papers mention MNIST?' returns all of documents that have mnist as a NER, a count of 50 documents. Similarly, 'Which documents in the security decomposition are in the 'cs.SE' (software engineering) arXiv category' finds 7 documents, which are further linked to 2 organizations, an event, topics 0 and 2, a product, and 7 additional categories.

\section{Conclusion}
\label{sec:conclusion}
In this paper, we have presented a concept for building a domain specific multi-model knowledge graph (KG) from unstructured scientific text data. Our KG includes observable entities from data such as paper titles, authors, publication year, as well as latent (hidden) patterns such as named entities, keywords, and topics/clusters that are extracted with a new semantic hierarchical NMF method, named HSNMFk-SPLIT, for document organization. HSNMFk-SPLIT is designed for large corpora using distributed joint factorization, and has the capability to perform automatic selection of the number of latent topics. The semantic structure and automatic model selection allows extraction of the coherent topics. We have demonstrated the feasibility of HSNMFk-SPLIT in performing topic and sub-topic extraction on the abstracts of more than two million papers posted on arXiv, and consolidating these generic scientific papers to cyber-specific data to build a highly-domain-specific KG.

\section*{Acknowledgment}
 This research was funded by the Los Alamos National Laboratory (LANL) Laboratory Directed Research and Development (LDRD) grant 20230067DR and  LANL Institutional Computing Program, supported by the U.S. Department of Energy National Nuclear Security Administration under Contract No. 89233218CNA000001.

\bibliographystyle{IEEEtran}
\bibliography{references.bib}

\end{document}